%% file: nips2015.tex
\documentclass[usenames,dvipsnames]{article}
\usepackage{nips15submit_e,times}

\input{preamble.tex}
\input{preamble_acronyms.tex}
\input{preamble_math.tex}

\usepackage{setspace}

\title{Copula variational inference}

\author{
Dustin Tran\\
Harvard University\\
\And
David M. Blei \\
Columbia University \\
\And
Edoardo M. Airoldi  \\
Harvard University \\
}

\nipsfinalcopy

\usepackage{multirow} 

\begin{document}

\maketitle

\begin{abstract}
  We develop a general variational inference method that preserves
  dependency among the latent variables. Our method uses copulas to
  augment the families of distributions used in mean-field and
  structured approximations. Copulas model the dependency that is not
  captured by the original variational distribution, and thus the
  augmented variational family guarantees better approximations to the
  posterior. With stochastic optimization, inference on the augmented
  distribution is scalable. Furthermore, our strategy is generic: it
  can be applied to any inference procedure that currently uses the
  mean-field or structured approach.
  Copula variational inference has many advantages: it reduces bias;
  it is less sensitive to local optima; it is less sensitive to
  hyperparameters; and it helps characterize and interpret the
  dependency among the latent variables.
\end{abstract}

\input{sec_introduction.tex}
\input{sec_background.tex}
\input{sec_cvi.tex}
\input{sec_experiments.tex}
\input{sec_conclusion.tex}

\subsubsection*{Acknowledgments}
We thank Luke Bornn, Robin Gong, and Alp Kucukelbir for their
insightful comments.
This work is supported by NSF IIS-0745520, IIS-1247664, IIS-1009542,
ONR N00014-11-1-0651, DARPA FA8750-14-2-0009, N66001-15-C-4032,
Facebook, Adobe, Amazon, and the John Templeton Foundation.

\bibliographystyle{apalike}
\small
\bibliography{nips2015}

{
}

\end{document}

%% file: preamble.tex
\usepackage[T1]{fontenc}
\usepackage{tgtermes}
\usepackage{amsmath}

\usepackage{amssymb}
\newcommand{\mathbold}[1]{\ensuremath{\boldsymbol{\mathbf{#1}}}}

\usepackage[english]{babel}
\usepackage[parfill]{parskip}
\usepackage{afterpage}
\usepackage{framed}
\usepackage{xspace}

{\endMakeFramed}

\usepackage{lineno}

\usepackage{ragged2e}

\newcounter{parcount}

\definecolor{shadecolor}{gray}{0.9}
\newcommand{\red}[1]{\textcolor{BrickRed}{#1}}

\newcommand{\green}[1]{\textcolor{OliveGreen}{#1}}
\newcommand{\blue}[1]{\textcolor{MidnightBlue}{#1}}

\usepackage{graphicx}
\usepackage[labelfont=bf]{caption}
\usepackage[format=hang]{subcaption}

\usepackage{booktabs, array}

\let\chapter\section
\usepackage[algoruled]{algorithm2e}
\usepackage{listings}
\usepackage{fancyvrb}
\fvset{fontsize=\small}

\usepackage[numbers]{natbib}

\usepackage[colorlinks,linktoc=all]{hyperref}
\usepackage[all]{hypcap}
\hypersetup{citecolor=Blue}
\hypersetup{linkcolor=MidnightBlue}
\hypersetup{urlcolor=MidnightBlue}

\newcommand{\myeqp}[1]{Eq.\ref{eq:#1}}

\newcommand{\myfig}[1]{Figure~\ref{fig:#1}}

\newcommand{\myalg}[1]{Algorithm~\ref{alg:#1}}

\usepackage
[acronym,smallcaps,nowarn,section,nonumberlist]{glossaries}
\glsdisablehyper{}

\lstdefinestyle{mystyle}{
    commentstyle=\color{OliveGreen},
    numberstyle=\tiny\color{black!60},
    stringstyle=\color{BrickRed},
    basicstyle=\ttfamily\scriptsize,
    breakatwhitespace=false,
    breaklines=true,
    captionpos=b,
    keepspaces=true,
    numbers=none,
    numbersep=5pt,
    showspaces=false,
    showstringspaces=false,
    showtabs=false,
    tabsize=2
}

%% file: preamble_acronyms.tex
\newacronym{ELBO}{elbo}{evidence lower bound}
\newacronym{KL}{kl}{Kullback-Leibler}

\newacronym{MF}{mf}{mean-field}
\newacronym{CVI}{copula vi}{copula variational inference}
\newacronym{VI}{vi}{variational inference}
\newacronym{LRVB}{lrvb}{linear response variational Bayes}

%% file: preamble_math.tex
\DeclareRobustCommand{\mb}[1]{\ensuremath{\boldsymbol{\mathbf{#1}}}}

\newcommand{\mbx}{\mathbold{x}}

\newcommand{\mbz}{\mathbold{z}}

\newcommand{\mblambda}{\mathbold{\lambda}}

\newcommand{\mbeta}{\mathbold{\eta}}

\newcommand{\mbmu}{\mathbold{\mu}}

\newcommand{\mbu}{\mathbold{u}}

\newcommand{\mbpi}{\mathbold{\pi}}
\newcommand{\mbLambda}{\mathbold{\Lambda}}
\newcommand{\g}{\,|\,}

%% file: sec_introduction.tex
\section{Introduction}
\label{sec:introduction}


Variational inference is a computationally efficient approach for
approximating posterior distributions. The idea is to specify a
tractable family of distributions of the latent variables and then to
minimize the Kullback-Leibler divergence  from it to the posterior. Combined
with stochastic optimization, variational inference can scale complex
statistical models to massive data sets~\citep{hoffman2013stochastic,toulis2014implicit,tran2015stochastic}.

Both the computational complexity and accuracy of variational
inference are controlled by the factorization of the variational
family. To keep optimization tractable, most algorithms use the
fully-factorized family, also known as the mean-field family, where each
latent variable is assumed independent. Less common, structured
mean-field methods slightly relax this assumption by preserving some
of the original structure among the latent
variables~\citep{saul1995exploiting}. Factorized distributions enable
efficient variational inference but they sacrifice accuracy. In the
exact posterior, many latent variables are dependent and mean-field
methods, by construction, fail to capture this dependency.


To this end, we develop \gls{CVI}.
\Gls{CVI} augments the traditional variational distribution with a
copula, which is a flexible construction for learning dependencies in
factorized distributions~\citep{frechet1960tableaux}.  This strategy has many advantages over
traditional \glsunset{VI}\gls{VI}: it reduces bias; it is less sensitive to local optima;
it is less sensitive to hyperparameters; and it helps characterize and
interpret the dependency among the latent variables.  Variational
inference has previously been restricted to either generic inference
on simple models---where dependency does not make a significant
difference---or writing model-specific variational updates. \Gls{CVI}
widens its applicability, providing generic inference that finds
meaningful dependencies between latent variables.

In more detail, our contributions are the following.

\begin{figure}[t]
\begin{minipage}{\textwidth}
  \begin{minipage}[b]{0.245\textwidth}
\includegraphics[width=\textwidth]{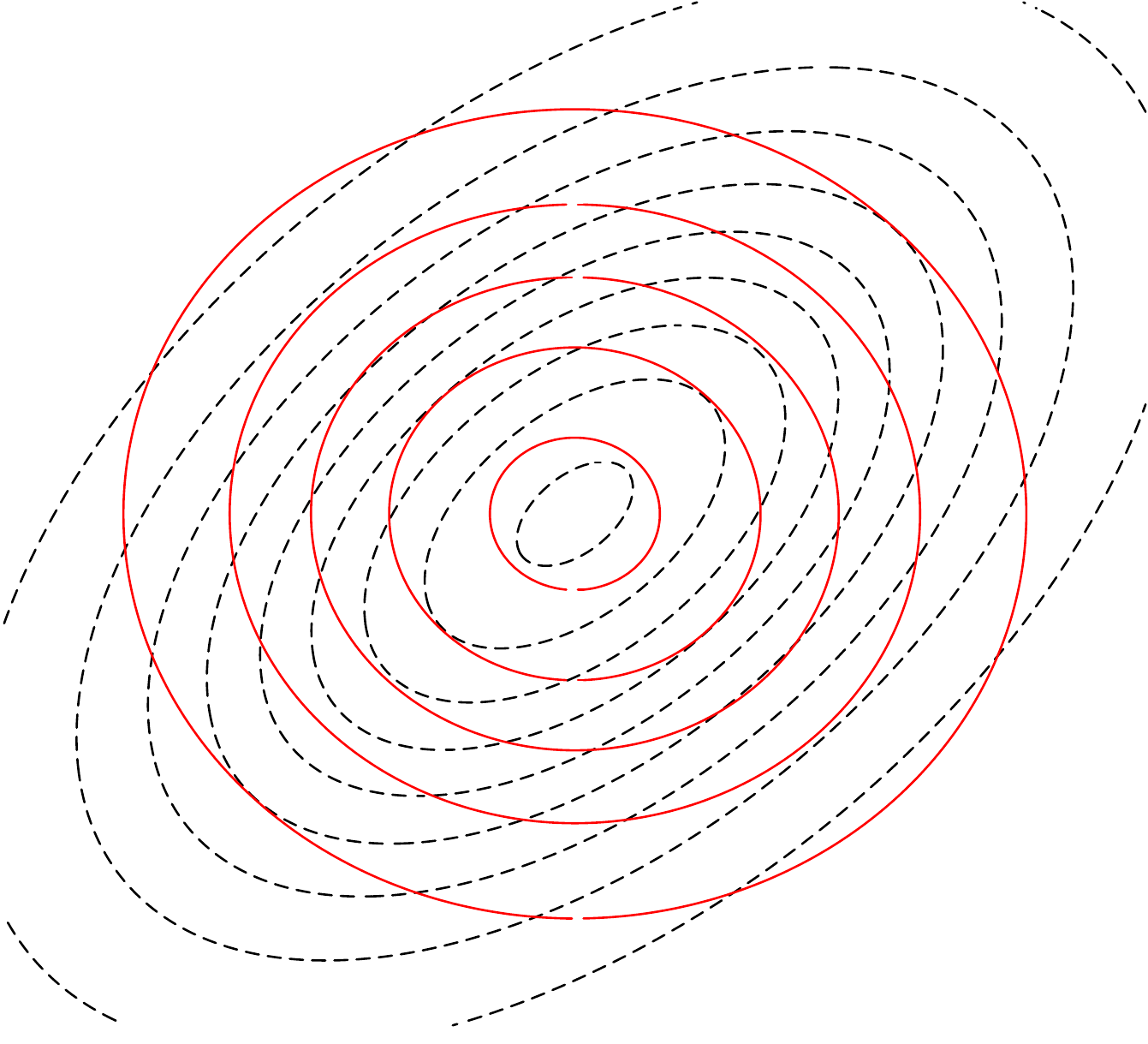}
  \end{minipage}
  \hfill
  \begin{minipage}[b]{0.245\textwidth}
\includegraphics[width=\textwidth]{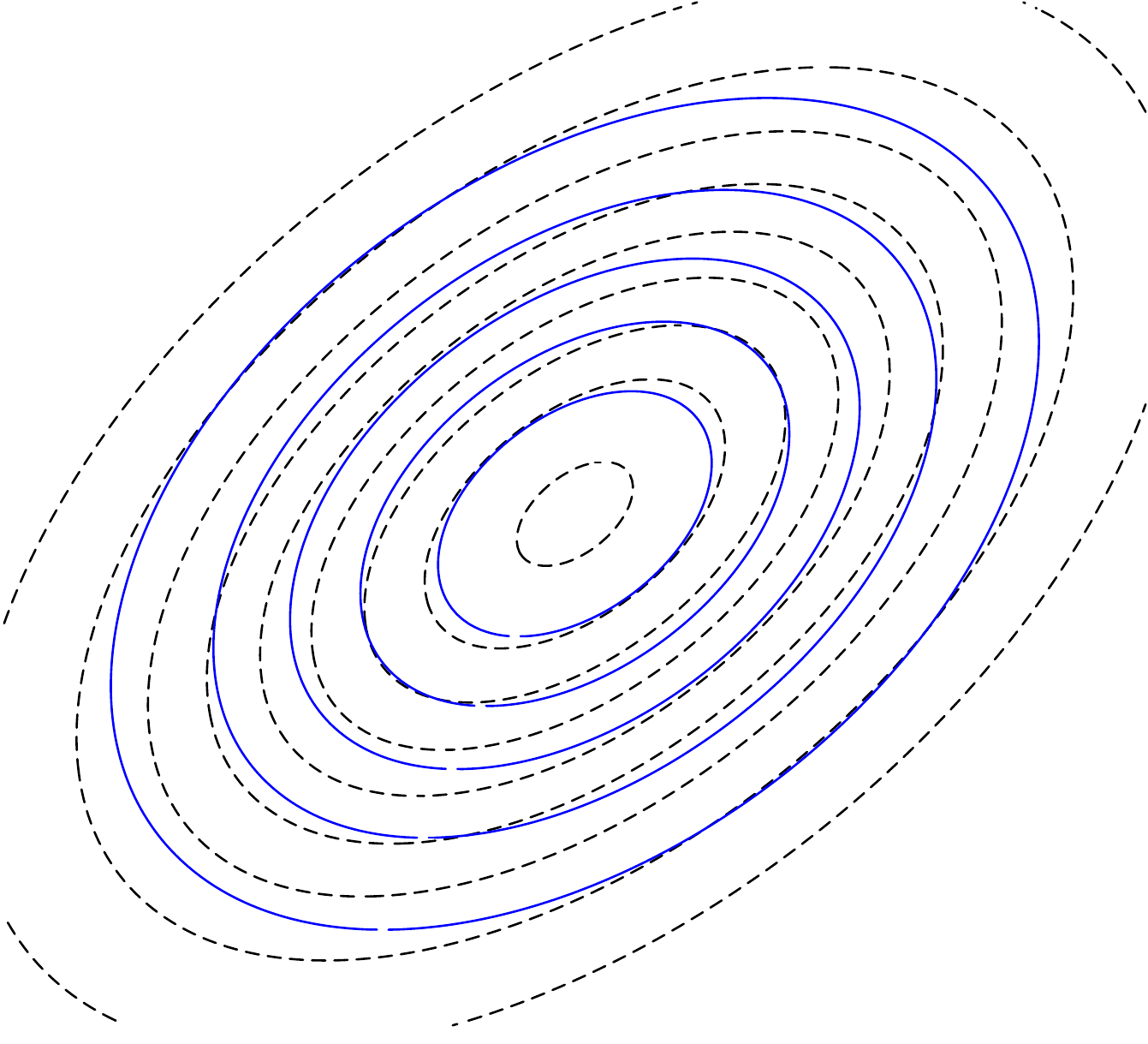}
  \end{minipage}
  \hfill
  \begin{minipage}[b]{0.245\textwidth}
\includegraphics[width=\textwidth]{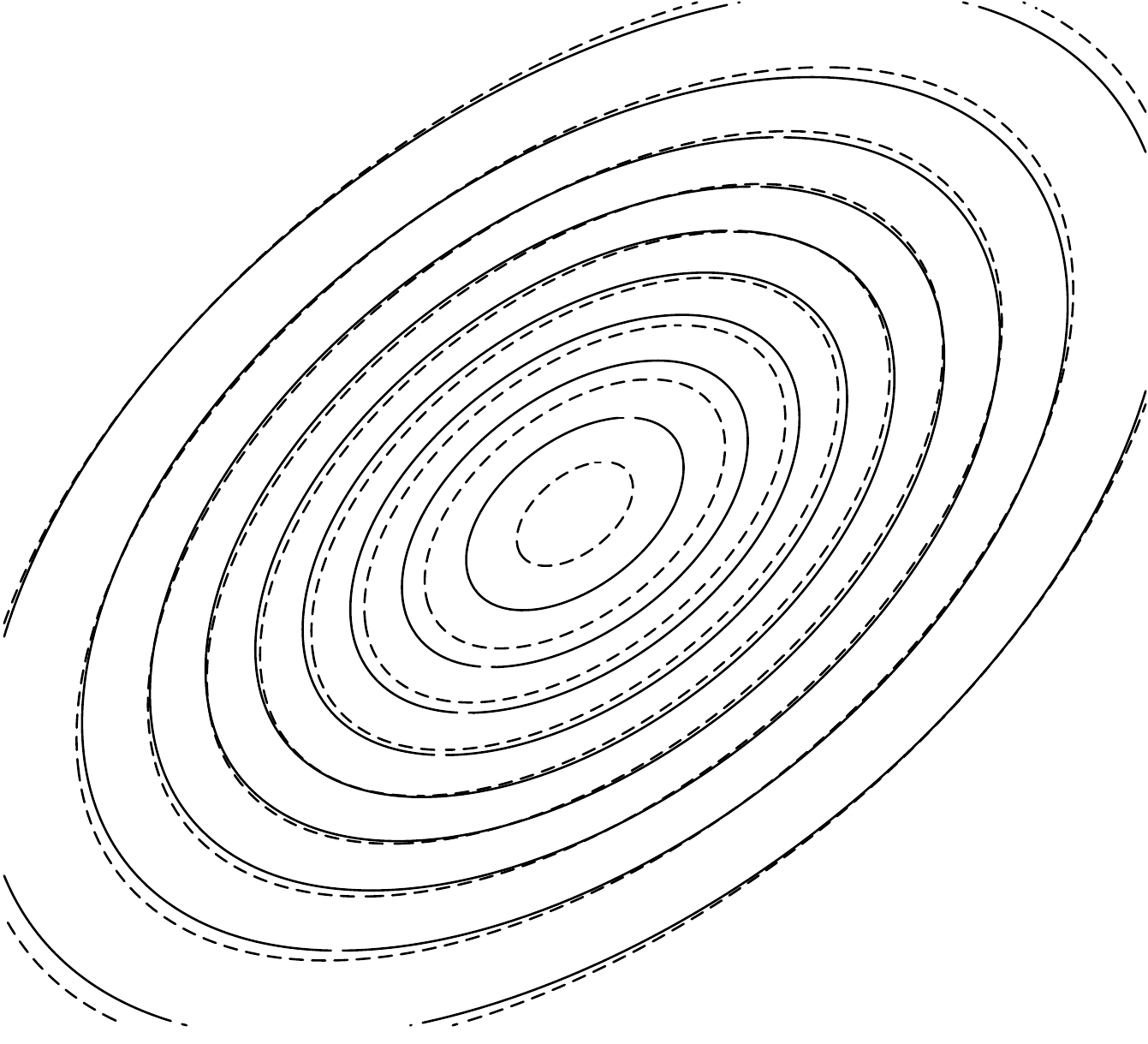}
  \end{minipage}
  \hfill
  \begin{minipage}[b]{0.245\textwidth}
\includegraphics[width=\textwidth]{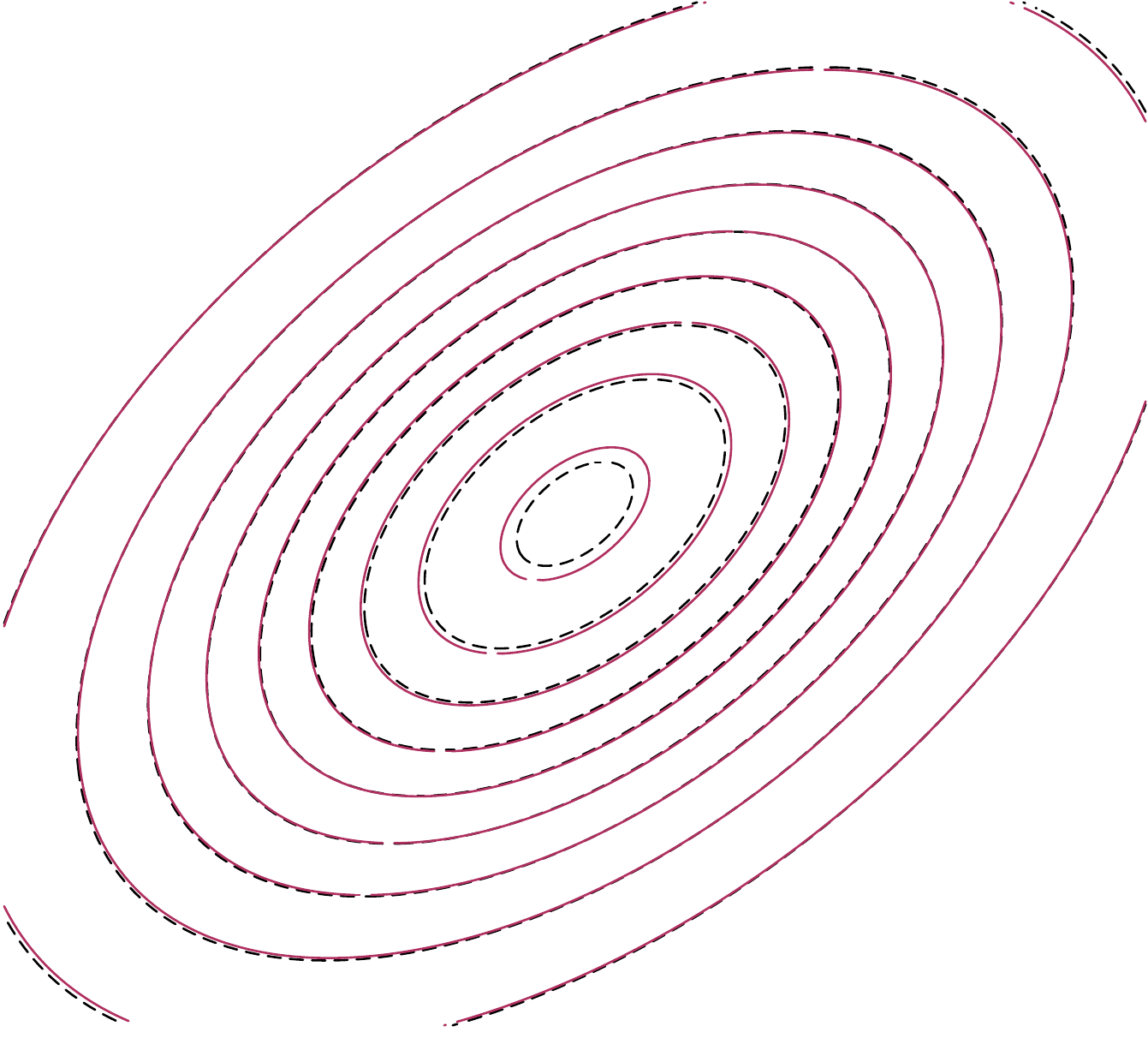}
  \end{minipage}
\end{minipage}
\caption{\label{fig:gaussian}Approximations to an
elliptical Gaussian. The mean-field (\textcolor{purple}{red}) is
restricted to fitting independent one-dimensional Gaussians, which is the first
step in our algorithm. The second step (\blue{blue}) fits a copula which models
the dependency. More iterations alternate: the third refits the mean-field
(\green{green}) and the fourth refits the copula
(\red{cyan}),
demonstrating convergence to the true posterior.}
\end{figure}

\textbf{A generalization of the original procedure in variational
  inference}. \Gls{CVI} generalizes variational inference for
mean-field and structured factorizations: traditional
\glsunset{VI}\gls{VI} corresponds
to running only one step of our method. It uses coordinate descent,
which monotonically decreases the KL divergence to the posterior by
alternating between fitting the mean-field parameters and the copula
parameters. Figure \ref{fig:gaussian} illustrates \gls{CVI} on a toy
example of fitting a bivariate Gaussian.

\textbf{Improving generic inference}.  \Gls{CVI} can be applied to any
inference procedure that currently uses the mean-field or structured
approach.  Further, because it does not require specific knowledge of
the model, it falls into the framework of black box variational
inference~\citep{ranganath2014black}.  An investigator need only write
down a function to evaluate the model log-likelihood. The rest of the
algorithm's calculations, such as sampling and evaluating gradients, can be placed in a library.

\textbf{Richer variational approximations}.  In experiments, we demonstrate
\gls{CVI} on the standard example of Gaussian mixture models.  We
found it consistently estimates the parameters, reduces sensitivity to
local optima, and reduces sensitivity to hyperparameters. We also
examine how well \gls{CVI} captures dependencies on the latent space
model~\citep{hoff2001latent}. \Gls{CVI} outperforms competing methods
and significantly improves upon the mean-field approximation.

%% file: sec_background.tex
\section{Background}
\label{sec:background}
\subsection{Variational inference}
\label{subsec:variational}

Let $\mbx$ be a set of observations, $\mbz$ be latent variables, and
$\mblambda$ be the free parameters of a variational distribution
$q(\mbz; \mblambda)$. We aim to find the best approximation of the
posterior $p(\mbz\g \mbx)$ using the variational distribution
$q(\mbz; \mblambda)$, where the quality of the approximation is
measured by KL divergence. This is equivalent to maximizing the
quantity
\begin{equation*}
\mathcal{L}\left(\mblambda\right)
 = \mathbb{E}_{q(\mbz;\mblambda)}[\log p(\mbx,\mbz)]
 - \mathbb{E}_{q(\mbz;\mblambda)}[\log
 q(\mbz;\mblambda)].
\end{equation*}
$\mathcal{L}(\mblambda)$ is the \emph{\gls{ELBO}}, or the variational free
energy~\citep{wainwright2008graphical}.
%
For simpler computation, a standard choice of the variational family
is a \emph{mean-field approximation}
\begin{equation*}
  q(\mbz;\mblambda) = \prod_{i=1}^{d} q_{i}(\mbz_{i};\mblambda_i),
\end{equation*}
where $\mbz=(\mb\mbz_1,\ldots,\mb\mbz_d)$. Note this is a strong independence
assumption.  More sophisticated approaches, known as \emph{structured
  variational inference}~\citep{saul1995exploiting}, attempt to restore
some of the dependencies among the latent variables.

In this work, we restore dependencies using copulas. Structured
\gls{VI} is typically tailored to individual models and is difficult
to work with mathematically. Copulas learn general posterior
dependencies during inference, and they do not require the
investigator to know such structure in advance. Further, copulas can
augment a structured factorization in order to introduce
dependencies that were not considered before; thus it generalizes
the procedure. We next review copulas.

\subsection{Copulas}
\label{subsec:copulas}

We will augment the mean-field distribution with a \emph{copula}. We
consider the variational family
\begin{equation*}
  q(\mbz) = \left[\prod_{i=1}^d q(\mbz_i)\right] c(Q(\mbz_1),\ldots,Q(\mbz_d)).
\end{equation*}
Here $Q(\mbz_i)$ is the marginal cumulative distribution function (CDF)
of the random variable $\mbz_i$, and $c$ is a joint distribution of
$[0,1]$ random variables.\footnote{We overload the notation for the
  marginal CDF $Q$ to depend on the names of the argument, though we
  occasionally use $Q_i(\mbz_i)$ when more clarity is needed. This is
  analogous to the standard convention of overloading the probability
  density function $q(\cdot)$.}  The distribution $c$ is called
a copula of $\mbz$: it is a joint multivariate density of
$Q(\mbz_1),\ldots,Q(\mbz_d)$ with uniform marginal
distributions~\citep{sklar1959fonstions}.  For any distribution, a
factorization into a product of marginal densities and a copula always
exists and integrates to one~\citep{nelsen2006introduction}.

Intuitively, the copula captures the information about the
multivariate random variable after eliminating the marginal
information, i.e., by applying the probability integral transform on
each variable. The copula captures only and all of the dependencies
among the $\mbz_i$'s. Recall that, for all random variables, $Q(\mbz_i)$ is
uniform distributed. Thus the marginals of the copula give no
information.

For example, the bivariate Gaussian copula is defined as
\begin{equation*}
  c(\mbu_1, \mbu_2; \rho)= \Phi_\rho(\Phi^{-1}(\mbu_1),
  \Phi^{-1}(\mbu_2)).
\end{equation*}
If $\mbu_1,\mbu_2$ are independent uniform distributed, the inverse CDF
$\Phi^{-1}$ of the standard normal transforms $(\mbu_1,\mbu_2)$ to independent
normals. The CDF $\Phi_{\rho}$ of the bivariate Gaussian
distribution, with mean zero and Pearson correlation $\rho$, squashes the
transformed values back to the unit square.
Thus the Gaussian copula directly correlates $\mbu_1$ and $\mbu_2$
with the Pearson correlation parameter $\rho$.

\subsubsection{Vine copulas}
\label{subsubsec:vine}

It is difficult to specify a copula.  We must find a family of
distributions that is easy to compute with and able to express a broad
range of dependencies.  Much work focuses on two-dimensional copulas,
such as the Student-$t$, Clayton, Gumbel, Frank, and Joe copulas
\citep{nelsen2006introduction}. However, their multivariate extensions
do not flexibly model dependencies in higher
dimensions~\citep{genest2009editorial}. Rather, a successful approach
in recent literature has been by combining sets of conditional
bivariate copulas; the resulting joint is called a \emph{vine}~\citep{joe1996families,kurowicka2006uncertainty}.

A vine $\mathcal{V}$ factorizes a copula density $c(\mbu_1,\ldots,\mbu_d)$
into a product of conditional bivariate copulas, also called pair
copulas.  This makes it easy to specify a high-dimensional copula.
One need only express the dependence for each pair of random variables
conditioned on a subset of the others.

\begin{figure}[t]
\centering
\includegraphics[width=0.75\textwidth]{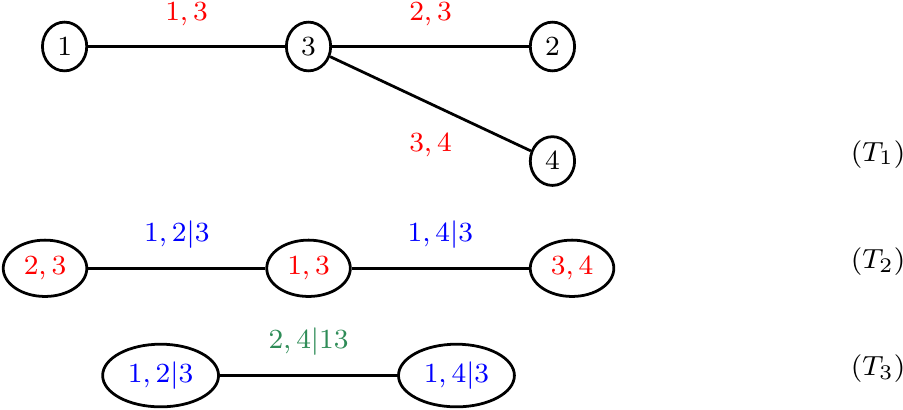}
\caption{Example of a vine $\mathcal{V}$ which factorizes a copula density of
four random variables $c(\mbu_1,\mbu_2,\mbu_3,\mbu_4)$ into a product
of 6 pair
copulas. Edges in the tree $T_j$ are the nodes of the lower level tree
$T_{j+1}$, and each edge determines a bivariate copula which is conditioned on
all random variables that its two connected nodes share.}
\label{fig:vine_copula}
\end{figure}

\myfig{vine_copula} is an example of a vine which factorizes a
4-dimensional copula into the product of 6 pair copulas. The first
tree $T_1$ has nodes $1,2,3,4$ representing the random variables
$\mbu_1,\mbu_2,\mbu_3,\mbu_4$ respectively. An edge corresponds
to a pair copula, e.g., $1,4$ symbolizes $c(\mbu_1,\mbu_4)$. Edges in
$T_1$ collapse into nodes in the next tree $T_2$, and edges in $T_2$
correspond to conditional bivariate copulas, e.g., $1,2|3$ symbolizes
$c(\mbu_1,\mbu_2|\mbu_3)$. This proceeds to the last
nested tree $T_3$, where $2,4|13$ symbolizes
$c(\mbu_2,\mbu_4|\mbu_1,\mbu_3)$. The vine structure specifies a
complete factorization of the multivariate copula, and each pair
copula can be of a different family with its own set of parameters:
\begin{align*}
c(\mbu_1,\mbu_2,\mbu_3,\mbu_4) =
\Big[c(\mbu_1,\mbu_3)c(\mbu_2,\mbu_3)c(\mbu_3,\mbu_4)\Big]
\Big[c(\mbu_1,\mbu_2|\mbu_3)c(\mbu_1,\mbu_4|\mbu_3)\Big]
\Big[c(\mbu_2,\mbu_4|\mbu_1,\mbu_3)\Big].
\end{align*}
Formally, a vine is a nested set of trees
$\mathcal{V} = \{T_1,\ldots, T_{d-1}\}$ with the following
properties:
\begin{enumerate}
\item Tree $T_j=\{N_j,E_j\}$ has $d+1-j$ nodes and $d-j$ edges.
\item Edges in the $j^{th}$ tree $E_j$ are the nodes in the $(j+1)^{th}$ tree $N_{j+1}$.
\item 
Two nodes in tree $T_{j+1}$ are joined by an edge only if the corresponding edges in tree $T_{j}$ share a node.
\end{enumerate}
Each edge $e$ in the nested set of trees $\{T_1,\ldots,T_{d-1}\}$
specifies a different pair copula, and the product of all edges
comprise of a factorization of the copula density. Since there are a
total of $d(d-1)/2$ edges, $\mathcal{V}$ factorizes
$c(\mbu_1,\ldots,\mbu_d)$ as the product of $d(d-1)/2$ pair copulas.

Each edge
$e(i,k) \in T_j$ has a
\emph{conditioning set} $D(e)$, which is a set of variable indices
$1,\ldots,d$. We define
$c_{ik|D(e)}$ to be the bivariate copula density for $\mbu_i$ and
$\mbu_k$ given its conditioning set:
\begin{equation}
c_{ik|D(e)} = c\Big( Q(\mbu_i|\mbu_j : j \in D(e)), Q(\mbu_i|\mbu_j :
j \in D(e)) \Big| \mb\mbu_j : j\in D(e)\Big).
\label{eq:defcjk}
\end{equation}
Both the copula and the CDF's in its arguments are conditional
on $D(e)$. A vine specifies a factorization of the
copula, which is a product over all edges in the $d-1$ levels:
\begin{equation}
  c(\mbu_1,\ldots,\mbu_d; \mbeta) = \prod_{j=1}^{d-1}\prod_{e(i,k)\in E_j} c_{ik|D(e)}.
\label{eq:vinedecomp}
\end{equation}
%
%
We highlight that $c$ depends on $\mbeta$, the set of all parameters to
the pair copulas.  The vine construction provides us with the
flexibility to model dependencies in high dimensions using a
decomposition of pair copulas which are easier to estimate. As we
shall see,
the construction also leads to efficient stochastic gradients by
taking individual (and thus easy) gradients on each pair copula.


%% file: sec_cvi.tex
\section{Copula variational inference}
\label{sec:copula}

We now introduce \glsreset{CVI}\emph{\gls{CVI}}, our method for
performing accurate and scalable variational inference. For
simplicity, consider the mean-field factorization augmented with a
copula (we later extend to structured factorizations). The copula-augmented variational family is
\begin{equation}
q(\mbz;\mblambda,\mbeta) = \underbrace{\left[\prod_{i=1}^d
q(\mbz_i;\mblambda)\right]}_{\text{mean-field}}
\underbrace{c(Q(\mbz_1;\mblambda),\ldots,Q(\mbz_d;\mblambda);
\mbeta)}_{\text{copula}},
\label{eq:copula_variational}
\end{equation}
where $\mblambda$ denotes the mean-field parameters and $\mbeta$ the
copula parameters. With this family, we maximize the augmented \gls{ELBO},
\begin{equation*}
  \mathcal{L}\left(\mblambda,\mbeta\right)
  = \mathbb{E}_{q(\mbz;\mblambda,\mbeta)}[\log p(\mbx,\mbz)] - \mathbb{E}_{q(\mbz;\mblambda,\mbeta)}[\log
  q(\mbz;\mblambda,\mbeta)].
\end{equation*}
\Gls{CVI} alternates between two steps: 1) fix the copula parameters
$\mbeta$ and solve for the mean-field parameters $\mblambda$; and 2)
fix the mean-field parameters $\mblambda$ and solve for the copula
parameters $\mbeta$. This generalizes the mean-field approximation,
which is the special case of initializing the copula to be uniform and
stopping after the first step. We apply stochastic approximations~\citep{robbins1951stochastic} for each step with
gradients derived in the next section. We set the learning rate
$\rho_t\in\mathbb{R}$ to satisfy a Robbins-Monro
schedule, i.e.,
$\sum_{t=1}^\infty\rho_t=\infty,~\sum_{t=1}^\infty\rho_t^2<\infty$.
A summary is outlined in \myalg{cvi}.

This alternating set of optimizations falls in the class of
minorize-maximization methods, which includes many procedures such as
the EM algorithm~\citep{dempster1777maximum}, the alternating least
squares algorithm, and the iterative procedure for the generalized
method of moments.
Each step of \gls{CVI}
monotonically increases the objective function and therefore better
approximates the posterior distribution.

\begin{algorithm}[t]
  \setstretch{0.1}
  \caption{\Acrfull{CVI}}
  \SetAlgoLined
  \DontPrintSemicolon
  \BlankLine
  \KwIn{Data $\mbx$, Model $p(\mbx,\mbz)$, Variational family $q$.}
  \BlankLine
  Initialize $\mblambda$ randomly, $\mbeta$ so that $c$ is uniform.\;
  \BlankLine
  \While{change in \gls{ELBO} is above some threshold}{
    \BlankLine
    // Fix $\mbeta$, maximize over $\mblambda$.\\[0.9ex]
    Set iteration counter $t = 1$.
    \BlankLine
    \While{not converged}{
      \BlankLine
      Draw sample $\mbu\sim \operatorname{Unif}([0,1]^d)$.
      \BlankLine
      Update
      $\mblambda = \mblambda + \rho_t \nabla_{\mblambda} \mathcal{L}$.
      (\myeqp{grad_meanfield:reinforce},
      \myeqp{grad_meanfield:reparam})
      \BlankLine
      Increment $t$.
      \BlankLine
    }
    \BlankLine
    // Fix $\mblambda$, maximize over $\mbeta$.\\[0.9ex]
    Set iteration counter $t = 1$.
    \BlankLine
    \While{not converged}{
      \BlankLine
      Draw sample $\mbu\sim \operatorname{Unif}([0,1]^d)$.
      \BlankLine
      Update
      $\mbeta = \mbeta + \rho_t \nabla_{\mbeta} \mathcal{L}$.
      (\myeqp{grad_copula})
      \BlankLine
      Increment $t$.
      \BlankLine
    }
    \BlankLine
  }
  \textbf{Output}: Variational parameters $(\mblambda,\mbeta)$.\;
  \label{alg:cvi}
\end{algorithm}
\Gls{CVI} has the same generic input requirements as black-box
variational inference~\citep{ranganath2014black}---the user need only
specify the joint model $p(\mbx,\mbz)$ in order to perform inference.
Further, \acrlong{CVI} easily extends to the case when the original
variational family uses a structured factorization. By the vine
construction, one simply fixes pair copulas corresponding to
pre-existent dependence in the factorization to be the independence
copula. This enables the copula to only model dependence where it does
not already exist.

Throughout the optimization, we will assume that the tree structure
and copula families are given and fixed.  We note, however, that these
can be learned. In our study, we learn the tree structure using
sequential tree selection~\citep{dissmann2012selecting} and learn the
families, among a choice of 16 bivariate families, through Bayesian
model selection~\citep{gruber2015sequential} (see supplement). In
preliminary studies, we've found that re-selection of the tree
structure and copula families do not significantly change in future
iterations.

\subsection{Stochastic gradients of the \gls{ELBO}}
\label{subsec:derivatives}

To perform stochastic optimization, we require stochastic gradients of
the \gls{ELBO} with respect to both the mean-field and copula
parameters. The \gls{CVI} objective leads to efficient stochastic
gradients and with low variance.

We first derive the gradient with respect to the mean-field
parameters.  In general, we can apply the score function
estimator~\citep{ranganath2014black}, which leads to the gradient
\begin{equation}
\nabla_{\mblambda}\mathcal{L} = \mathbb{E}_{q(\mbz;\mblambda,\mbeta)}[\nabla_{\mblambda}\log
q(\mbz;\mblambda,\mbeta)\cdot(\log p(\mbx,\mbz)-\log
q(\mbz;\mblambda,\mbeta))].
\label{eq:grad_elbo}
\end{equation}
We follow noisy unbiased estimates of this gradient by sampling from
$q(\cdot)$ and evaluating the inner expression.  We apply this gradient for
discrete latent variables.

When the latent variables $\mbz$ are differentiable, we use the
reparameterization trick~\citep{rezende2014stochastic} to take advantage of first-order
information from the model, i.e.,$\nabla_{\mbz} \log p(\mbx,\mbz)$.
Specifically, we rewrite the expectation in terms of a random variable
$\mbu$ such that its distribution $s(\mbu)$ does not depend on the
variational parameters and such that the latent variables are a
deterministic function of $\mbu$ and the mean-field parameters,
$\mbz=\mbz(\mbu;\mblambda)$. Following this reparameterization, the
gradients propagate inside the expectation,
\begin{equation}
  \nabla_{\mblambda}\mathcal{L} =
  \mathbb{E}_{s(\mbu)}[(\nabla_{\mbz}\log
  p(\mbx,\mbz)-\nabla_{\mbz}\log
  q(\mbz;\mblambda,\mbeta))\nabla_{\mblambda}\mbz(\mbu;\mblambda)].
\label{eq:grad_meanfield:reinforce}
\end{equation}
This estimator reduces the variance of the stochastic
gradients~\citep{rezende2014stochastic}.
Furthermore, with a copula variational family, this type of
reparameterization using a uniform random variable $\mbu$ and a deterministic function $\mbz=\mbz(\mbu;\mblambda,\mbeta)$ is always possible. (See the supplement.)

The reparameterized gradient (\myeqp{grad_meanfield:reinforce})
requires calculation of the terms
$\nabla_{\mbz_i}\log q(\mbz ; \mblambda,\mbeta)$ and
$\nabla_{\mblambda_i}\mbz(\mbu;\mblambda,\mbeta)$ for each $i$. The latter is tractable and derived in the supplement; the former decomposes as
\begin{align}
  \nabla_{\mbz_i}\log q(\mbz; \mblambda, \mbeta)
                 &=
                   \nabla_{\mbz_i} \log q(\mbz_i; \mblambda_i) +
                   \nabla_{Q(\mbz_i;\mblambda_i)} \log c(Q(\mbz_1;\mblambda_1), \ldots, Q(\mbz_d; \mblambda_d); \mbeta)
                   \nabla_{\mbz_i} Q(\mbz_i;\mblambda_i)
                   \nonumber
  \\
                 &=
                   \nabla_{\mbz_i} \log q(\mbz_i; \mblambda_i)+
                   q(\mbz_i;\mblambda_i)
                   \sum_{j=1}^{d-1}\sum_{\substack{e(k,\ell)\in E_j:\\i\in
  \{k,\ell\}}}
  \nabla_{Q(\mbz_i;\mblambda_i)}
  \log c_{k\ell|D(e)}.
\label{eq:grad_meanfield:reparam}
\end{align}
The summation in \myeqp{grad_meanfield:reparam} is over all pair
copulas which contain $Q(\mbz_i ; \mblambda_i)$ as an argument. In
other words, the gradient of a latent variable $\mbz_i$ is evaluated
over both the marginal $q(\mbz_i)$ and all pair copulas which model
correlation between $\mbz_i$ and any other latent variable
$\mbz_j$.
A similar derivation holds for calculating terms in the score function estimator.

We now turn to the gradient with respect to the copula parameters. We
consider copulas which are differentiable with respect to their
parameters. This enables an efficient reparameterized gradient
\begin{equation}
\nabla_{\mbeta}\mathcal{L} =
\mathbb{E}_{s(\mbu)}[(\nabla_{\mbz}\log
p(\mbx,\mbz)-\nabla_{\mbz}\log
q(\mbz;\mblambda,\mbeta))\nabla_{\mbeta}\mbz(\mbu;\mblambda,\mbeta)].
\label{eq:grad_copula}
\end{equation}
The requirements are the same as for the mean-field parameters.

Finally, we note that the only requirement on the model is the
gradient $\nabla_{\mbz}\log p(\mbx,\mbz)$. This can be calculated
using automatic differentiation tools~\citep{stan-software:2015}. Thus \Gls{CVI} can be implemented in a library
and applied without requiring any manual derivations from the user.

\subsection{Computational complexity}
In the vine factorization of the copula, there are
$d(d-1)/2$ pair copulas, where $d$ is the number of
latent variables. Thus stochastic gradients of the
mean-field parameters $\mblambda$ and copula parameters $\mbeta$
require $\mathcal{O}(d^2)$ complexity.
More generally, one can apply a low rank approximation to the copula
by truncating the number of levels in the vine (see
\myfig{vine_copula}). This reduces the number of pair copulas to be
$Kd$ for some $K>0$, and leads to a computational complexity of
$\mathcal{O}(Kd)$.

Using sequential tree selection for learning the vine
structure~\citep{dissmann2012selecting}, the most correlated variables
are at the highest level of the vines. Thus a truncated low rank
copula only forgets the weakest correlations.  This generalizes low
rank Gaussian approximations, which also have $\mathcal{O}(Kd)$
complexity~\citep{seeger2010gaussian}: it is the special case when the
mean-field distribution is the product of independent Gaussians, and
each pair copula is a Gaussian copula.

\subsection{Related work}
Preserving structure in variational inference was first studied
by~\citet{saul1995exploiting} in the case of probabilistic neural
networks. It has been revisited recently for the case of conditionally
conjugate exponential familes~\citep{hoffman2015structured}. Our work
differs from this line in that we learn the dependency
structure during inference, and thus we do not require explicit
knowledge of the model. Further, our augmentation strategy works more
broadly to any posterior distribution and any factorized variational
family, and thus it generalizes these approaches.

A similar augmentation strategy is higher-order mean-field
methods,
which are a Taylor series correction
based on the difference between the posterior and its mean-field
approximation~\citep{kappen2001second}. Recently,
\citet{giordano2015linear} consider a covariance correction from the
mean-field estimates. All these methods assume the mean-field
approximation is reliable for the Taylor series expansion to make
sense, which is not true in general and thus is not robust in a black
box framework. Our approach alternates the estimation of the
mean-field and copula, which we find empirically leads to more robust
estimates than estimating them simultaneously, and which is
less sensitive to the quality of the mean-field approximation.


%% file: sec_experiments.tex
\section{Experiments}
\label{sec:experiments}

We study \gls{CVI} with two models: Gaussian mixtures and the latent
space model~\citep{hoff2001latent}. The Gaussian mixture is a
classical example of a model for which it is difficult to capture
posterior dependencies. The latent space model is a modern Bayesian
model for which the mean-field approximation gives poor estimates of
the posterior, and where modeling posterior dependencies is crucial for uncovering patterns in the data.

There are several implementation details of \gls{CVI}. At each
iteration, we form a stochastic gradient by generating $m$ samples
from the variational distribution and taking the average gradient. We
set $m=1024$ and follow asynchronous updates~\citep{recht2011hogwild}. We set the step-size using ADAM~\citep{kingma2015adam}.

\subsection{Mixture of Gaussians}
\label{subsec:mixture}
We follow the goal of \citet{giordano2015linear}, which is to estimate
the posterior covariance for a Gaussian mixture. The hidden variables
are a $K$-vector of mixture proportions $\mbpi$ and a set of $K$
$P$-dimensional multivariate normals
$\mathcal{N}(\mbmu_k,\mbLambda_k^{-1})$, each with unknown mean
$\mbmu_k$ (a $P$-vector) and $P\times P$ precision matrix
$\mbLambda_k$. In a mixture of Gaussians, the joint probability is
\begin{align*}
p(\mbx, \mbz, \mbmu, \mbLambda^{-1},\mbpi)
&= p(\mbpi)\prod_{k=1}^Kp(\mbmu_k,\mbLambda^{-1}_k)\prod_{n=1}^N p(\mbx_n\g
\mbz_n,\mbmu_{\mbz_n},\mbLambda^{-1}_{\mbz_n})p(\mbz_n\g \mbpi),
\end{align*}
with a Dirichlet prior $p(\mbpi)$ and a normal-Wishart prior
$p(\mbmu_k,\mbLambda_k^{-1})$.

We first apply the mean-field approximation (\glsunset{MF}\gls{MF}),
which assigns independent factors to $\mbmu, \mbpi, \mbLambda$, and
$\mbz$. We then perform \gls{CVI} over the copula-augmented
mean-field distribution, i.e., one which includes pair copulas over
the latent variables. We also compare our results to
\gls{LRVB}~\citep{giordano2015linear}, which is a posthoc correction
technique for covariance estimation in variational inference.
Higher-order mean-field methods demonstrate similar behavior as
\gls{LRVB}. Comparisons to structured approximations are omitted as
they require explicit factorizations and are not black box. Standard black box variational inference~\citep{ranganath2014black}
corresponds to the \gls{MF} approximation.

\begin{figure}[t]
  \centering
  \includegraphics[width=0.7\textwidth]{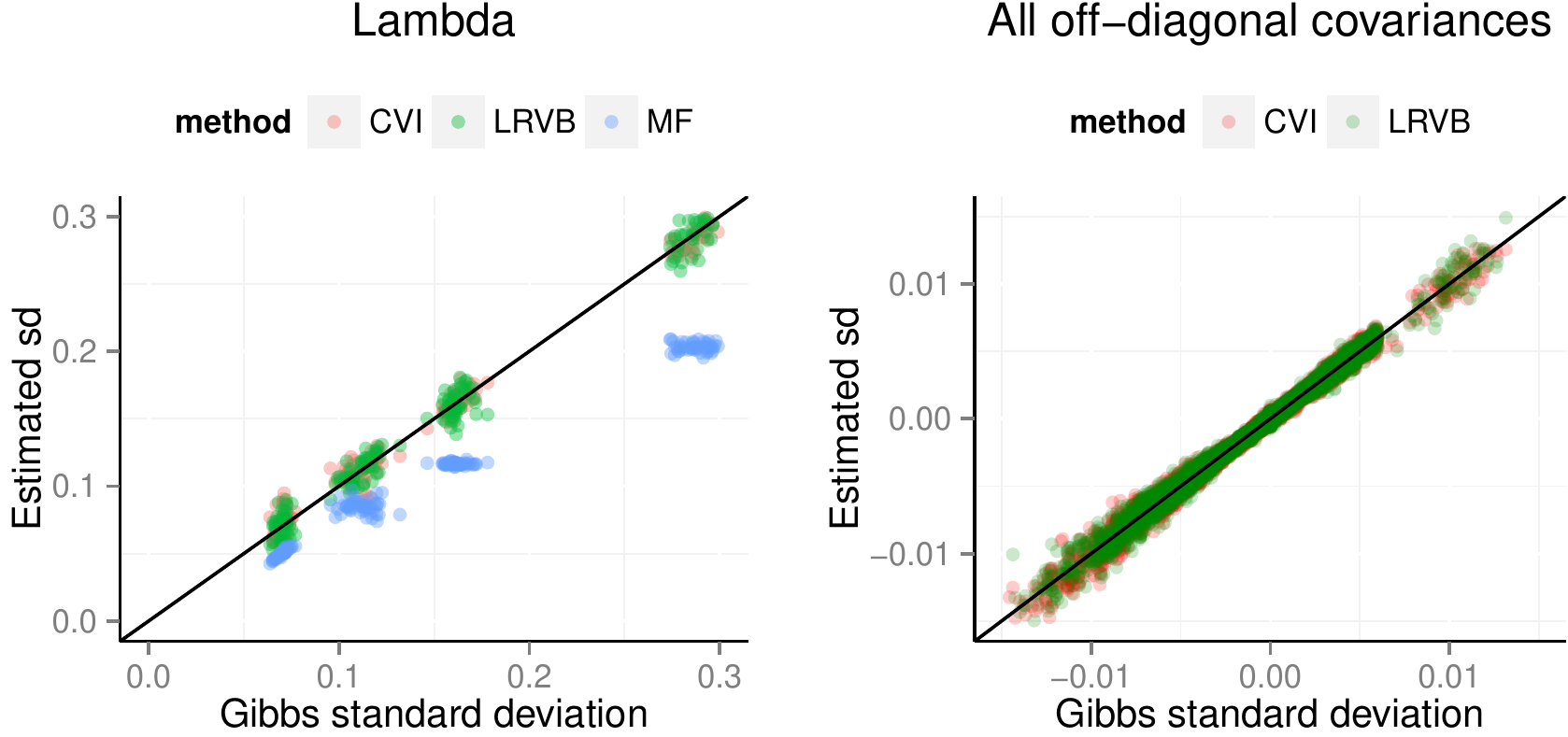}
  \caption{\label{fig:lrvb}Covariance estimates from
  copula variational inference (\gls{CVI}), mean-field (\gls{MF}), and
  linear response variational Bayes (\gls{LRVB}) to the ground truth
  (Gibbs samples). \gls{CVI} and \gls{LRVB} effectively capture dependence
  while \gls{MF} underestimates variance and forgets covariances.}
\end{figure}

We simulate $10,000$ samples with $K=2$ components and $P=2$
dimensional Gaussians. Figure \ref{fig:lrvb} displays estimates for
the standard deviations of $\mbLambda$ for 100 simulations, and plots
them against the ground truth using 500 effective Gibb samples. The
second plot displays all off-diagonal covariance estimates. Estimates
for $\mbmu$ and $\mbpi$ indicate the same pattern and are given in the
supplement.

When initializing at the true mean-field parameters, both \gls{CVI}
and \gls{LRVB} achieve consistent estimates of the posterior variance. \gls{MF} underestimates the variance, which is a well-known
limitation~\citep{wainwright2008graphical}. Note that
because the \gls{MF} estimates are initialized at the truth, \gls{CVI}
converges to the true posterior upon one step of fitting the copula. It does not require alternating more steps.

\Gls{CVI} is more robust than \gls{LRVB}. As a toy demonstration, we
analyze the MNIST data set of handwritten digits, using 12,665
training examples and 2,115 test examples of 0's and 1's. We perform
"unsupervised" classification, i.e., classify without using training
labels: we apply a mixture of Gaussians to cluster, and then classify
a digit based on its membership assignment. \gls{CVI} reports a test
set error rate of 0.06, whereas \gls{LRVB} ranges between 0.06 and
0.32 depending on the mean-field estimates. \gls{LRVB} and similar
higher order mean-field methods correct an existing \gls{MF}
solution---it is thus sensitive to local optima and the general
quality of that solution. On the other hand, \gls{CVI} re-adjusts both
the \gls{MF} and copula parameters as it fits, making it more robust
to initialization.

\subsection{Latent space model}
\label{subsec:latent}

We next study inference on the latent space
model~\citep{hoff2001latent}, a Bernoulli latent factor model for
network analysis. Each node in an $N$-node network is associated with
a $P$-dimensional latent variable $\mbz\sim N(\mbmu,\mbLambda^{-1})$.
Edges between pairs of nodes are observed with high probability if the
nodes are close to each other in the latent space. Formally, an edge
for each pair $(i,j)$ is observed with probability $\mathrm{logit}(p)=
\theta - |\mbz_i - \mbz_j|$, where $\theta$ is a model parameter.

We generate an $N=100,000$ node network with latent node attributes
from a $P=10$ dimensional Gaussian. We learn the posterior of the
latent attributes in order to predict the likelihood of held-out
edges. \gls{MF} applies independent factors on $\mbmu,
\mbLambda,\theta$ and $\mbz$, \gls{LRVB} applies a correction, and
\gls{CVI} uses the fully dependent variational distribution. Table
\ref{table:latent} displays the likelihood of held-out edges and runtime. We also attempted Hamiltonian Monte Carlo but it did not
converge after five hours.

\Gls{CVI} dominates other methods in accuracy upon convergence, and
the copula estimation without refitting (2 steps) already dominates
\gls{LRVB} in both runtime and accuracy. We note however that
\gls{LRVB} requires one to invert a
$\mathcal{O}(NK^3)\times\mathcal{O}(NK^3)$ matrix. We can better scale
the method and achieve faster estimates than \gls{CVI} if we applied
stochastic approximations for the inversion. However, \gls{CVI} always
outperforms \gls{LRVB} and is still fast on this 100,000 node network.

\begin{table}[t]
  \centering
  \begin{tabular}{lll}
  \toprule
  Variational inference methods & Predictive Likelihood & Runtime\\
  \midrule
  {Mean-field} & -383.2 & 15 min.\\
  \gls{LRVB} & -330.5 & 38 min.\\
  \gls{CVI} (2 steps) & -303.2 & 32 min.\\
  \gls{CVI} (5 steps) & -80.2 & 1 hr. 17 min.\\
  \gls{CVI} (converged) & -50.5 & 2 hr.\\
  \bottomrule
  \end{tabular}
  \captionof{table}{\label{table:latent}Predictive likelihood on the
  latent space model. Each \gls{CVI} step either refits the mean-field
  or the copula. \gls{CVI} converges in roughly 10 steps and already
  significantly outperforms both mean-field and \gls{LRVB} upon
  fitting the copula once (2 steps).}
\end{table}


%% file: sec_conclusion.tex
\section{Conclusion}
\label{sec:conclusion}

We developed \glsreset{CVI}\gls{CVI}.  \gls{CVI} is a new variational
inference algorithm that augments the mean-field variational
distribution with a copula; it captures posterior dependencies among
the latent variables.  We derived a scalable and generic algorithm for
performing inference with this expressive variational distribution.
We found that \gls{CVI} significantly reduces the bias of the
mean-field approximation, better estimates the posterior variance, and
is more accurate than other forms of capturing posterior dependency in
variational approximations.


%% file: nips2015.bbl
\begin{thebibliography}{}

\bibitem[Dempster et~al., 1977]{dempster1777maximum}
Dempster, A.~P., Laird, N.~M., and Rubin, D.~B. (1977).
\newblock Maximum likelihood from incomplete data via the em algorithm.
\newblock {\em Journal of the Royal Statistical Society, Series B}, 39(1).

\bibitem[Dissmann et~al., 2012]{dissmann2012selecting}
Dissmann, J., Brechmann, E.~C., Czado, C., and Kurowicka, D. (2012).
\newblock Selecting and estimating regular vine copulae and application to
  financial returns.
\newblock {\em arXiv preprint arXiv:1202.2002}.

\bibitem[Fr{\'e}chet, 1960]{frechet1960tableaux}
Fr{\'e}chet, M. (1960).
\newblock Les tableaux dont les marges sont donn{\'e}es.
\newblock {\em Trabajos de estad{\'\i}stica}, 11(1):3--18.

\bibitem[Genest et~al., 2009]{genest2009editorial}
Genest, C., Gerber, H.~U., Goovaerts, M.~J., and Laeven, R. (2009).
\newblock Editorial to the special issue on modeling and measurement of
  multivariate risk in insurance and finance.
\newblock {\em Insurance: Mathematics and Economics}, 44(2):143--145.

\bibitem[Giordano et~al., 2015]{giordano2015linear}
Giordano, R., Broderick, T., and Jordan, M.~I. (2015).
\newblock Linear response methods for accurate covariance estimates from mean
  field variational bayes.
\newblock In {\em Neural Information Processing Systems}.

\bibitem[Gruber and Czado, 2015]{gruber2015sequential}
Gruber, L. and Czado, C. (2015).
\newblock Sequential bayesian model selection of regular vine copulas.
\newblock {\em International Society for Bayesian Analysis}.

\bibitem[Hoff et~al., 2001]{hoff2001latent}
Hoff, P.~D., Raftery, A.~E., and Handcock, M.~S. (2001).
\newblock Latent space approaches to social network analysis.
\newblock {\em Journal of the American Statistical Association}, 97:1090--1098.

\bibitem[Hoffman and Blei, 2015]{hoffman2015structured}
Hoffman, M.~D. and Blei, D.~M. (2015).
\newblock {Structured Stochastic Variational Inference}.
\newblock In {\em Artificial Intelligence and Statistics}.

\bibitem[Hoffman et~al., 2013]{hoffman2013stochastic}
Hoffman, M.~D., Blei, D.~M., Wang, C., and Paisley, J. (2013).
\newblock Stochastic variational inference.
\newblock {\em Journal of Machine Learning Research}, 14:1303--1347.

\bibitem[Joe, 1996]{joe1996families}
Joe, H. (1996).
\newblock {\em Families of $m$-variate distributions with given margins and
  $m(m-1)/2$ bivariate dependence parameters}, pages 120--141.
\newblock Institute of Mathematical Statistics.

\bibitem[Kappen and Wiegerinck, 2001]{kappen2001second}
Kappen, H.~J. and Wiegerinck, W. (2001).
\newblock {Second order approximations for probability models}.
\newblock In {\em Neural Information Processing Systems}.

\bibitem[Kingma and Ba, 2015]{kingma2015adam}
Kingma, D.~P. and Ba, J.~L. (2015).
\newblock {Adam: a Method for Stochastic Optimization}.
\newblock In {\em International Conference on Learning Representations}.

\bibitem[Kurowicka and Cooke, 2006]{kurowicka2006uncertainty}
Kurowicka, D. and Cooke, R.~M. (2006).
\newblock {\em Uncertainty Analysis with High Dimensional Dependence
  Modelling}.
\newblock Wiley, New York.

\bibitem[Nelsen, 2006]{nelsen2006introduction}
Nelsen, R.~B. (2006).
\newblock {\em An Introduction to Copulas (Springer Series in Statistics)}.
\newblock Springer-Verlag New York, Inc.

\bibitem[Ranganath et~al., 2014]{ranganath2014black}
Ranganath, R., Gerrish, S., and Blei, D.~M. (2014).
\newblock Black box variational inference.
\newblock In {\em Artificial Intelligence and Statistics}, pages 814--822.

\bibitem[Recht et~al., 2011]{recht2011hogwild}
Recht, B., Re, C., Wright, S., and Niu, F. (2011).
\newblock Hogwild: A lock-free approach to parallelizing stochastic gradient
  descent.
\newblock In {\em Advances in Neural Information Processing Systems}, pages
  693--701.

\bibitem[Rezende et~al., 2014]{rezende2014stochastic}
Rezende, D.~J., Mohamed, S., and Wierstra, D. (2014).
\newblock {Stochastic Backpropagation and Approximate Inference in Deep
  Generative Models}.
\newblock In {\em International Conference on Machine Learning}.

\bibitem[Robbins and Monro, 1951]{robbins1951stochastic}
Robbins, H. and Monro, S. (1951).
\newblock A stochastic approximation method.
\newblock {\em The Annals of Mathematical Statistics}, 22(3):400--407.

\bibitem[Saul and Jordan, 1995]{saul1995exploiting}
Saul, L. and Jordan, M.~I. (1995).
\newblock Exploiting tractable substructures in intractable networks.
\newblock In {\em Neural Information Processing Systems}, pages 486--492.

\bibitem[Seeger, 2010]{seeger2010gaussian}
Seeger, M. (2010).
\newblock Gaussian covariance and scalable variational inference.
\newblock In {\em International Conference on Machine Learning}.

\bibitem[Sklar, 1959]{sklar1959fonstions}
Sklar, A. (1959).
\newblock Fonstions de r\'epartition \`a $n$ dimensions et leurs marges.
\newblock {\em Publications de l'Institut de Statistique de l'Universit\'e de
  {P}aris}, 8:229--231.

\bibitem[{Stan Development Team}, 2015]{stan-software:2015}
{Stan Development Team} (2015).
\newblock Stan: A c++ library for probability and sampling, version 2.8.0.

\bibitem[Toulis and Airoldi, 2014]{toulis2014implicit}
Toulis, P. and Airoldi, E.~M. (2014).
\newblock {Implicit stochastic gradient descent}.
\newblock {\em arXiv preprint arXiv:1408.2923}.

\bibitem[Tran et~al., 2015]{tran2015stochastic}
Tran, D., Toulis, P., and Airoldi, E.~M. (2015).
\newblock {Stochastic gradient descent methods for estimation with large data
  sets}.
\newblock {\em arXiv preprint arXiv:1509.06459}.

\bibitem[Wainwright and Jordan, 2008]{wainwright2008graphical}
Wainwright, M.~J. and Jordan, M.~I. (2008).
\newblock {Graphical Models, Exponential Families, and Variational Inference}.
\newblock {\em Foundations and Trends in Machine Learning}, 1(1-2):1--305.

\end{thebibliography}
